\documentclass[letterpaper, 10 pt, conference]{ieeeconf}
\usepackage{graphicx}

\IEEEoverridecommandlockouts                              

\title{\LARGE \bf
The AdobeIndoorNav Dataset: Towards Deep Reinforcement Learning based Real-world Indoor Robot Visual Navigation
}

\author{Kaichun Mo, Haoxiang Li, Zhe Lin and Joon-Young Lee
\thanks{*Kaichun Mo is with Stanford University. This work is done while Kaichun Mo is an research intern at Adobe Research. Haoxiang Li, Zhe Lin and Joon-Young Lee are with Adobe Research.}
}

\begin{document}

\maketitle
\thispagestyle{empty}
\pagestyle{empty}

\begin{abstract}
Deep reinforcement learning (DRL) demonstrates its potential in learning a model-free navigation policy for robot visual navigation. However, the data-demanding algorithm relies on a large number of navigation trajectories in training. Existing datasets supporting training such robot navigation algorithms consist of either 3D synthetic scenes or reconstructed scenes. Synthetic data suffers from domain gap to the real-world scenes while visual inputs rendered from 3D reconstructed scenes have undesired holes and artifacts. In this paper, we present a new dataset collected in real-world to facilitate the research in DRL based visual navigation. Our dataset includes 3D reconstruction for real-world scenes as well as densely captured real 2D images from the scenes. It provides high-quality visual inputs with real-world scene complexity to the robot at dense grid locations. We further study and benchmark one recent DRL based navigation algorithm \cite{zhu2017target} and present our attempts and thoughts on improving its generalizability to unseen test targets in the scenes. 
\end{abstract}

\section{Introduction}



Autonomous navigation with visual sensors in an unknown environment is a fundamental and challenging topic in robotics research. Existing practical solution generally demands building a map of the environment with visual Simultaneous localization and mapping (SLAM) algorithms and planing navigation routes to move the robot from start locations to the target locations. While the state-of-the-art SLAM systems demonstrate strong navigation performance in the indoor environments, the recently proposed deep reinforcement learning (DRL) based methods~\cite{zhu2017target,piotr2016nav,zhang2016deep,tai2017virtual} open up a new possibility that avoids the complexity of constructing these sophisticated SLAM pipelines and directly learns model-free policies for robot visual navigation in a trial-and-error manner.

The success of deep learning is powered by big data~\cite{deng2009imagenet}. DRL based navigation needs a large number of trajectories in training. Unfortunately, it is not feasible to train such navigation methods online with real robots walking in the real environments due to the low data efficiency of current DRL algorithms. Instead, typical paradigm in training DRL is running a simulator to generate visual inputs to the robot and allow it to navigate in the virtual scenes. Such simulation demands real-world 3D scenes datasets that provide real visual inputs at every possible location that the robot can reach.

Existing 3D scenes datasets supporting the training of robot navigation can generally be categorized into two categories: 1) synthetic scenes~\cite{fisher2012example, song2016ssc, zhang2017physically, handa2015scenenet, mccormac2016scenenet, kolve2017ai2} and 2) 3D reconstructed scenes~\cite{chang2017matterport3d, armeni2017joint, dai2017scannet}. For example, the AI2-THOR framework~\cite{kolve2017ai2} proposed by Zhu et al.~\cite{zhu2017target} contains 120 synthetic scenes with photo-realistic textures. The Stanford 2D-3D-S~\cite{armeni2017joint}, ScanNet~\cite{dai2017scannet} and Matterport3D~\cite{chang2017matterport3d} datasets provide 3D real-world scenes with raw 3D point clouds and reconstructed meshes. They also collect sparsely sampled 2D observations and other semantic annotations to enable research in multi-modal scene understanding.

\begin{figure*}[t]
    \centering
    \includegraphics[width=\linewidth]{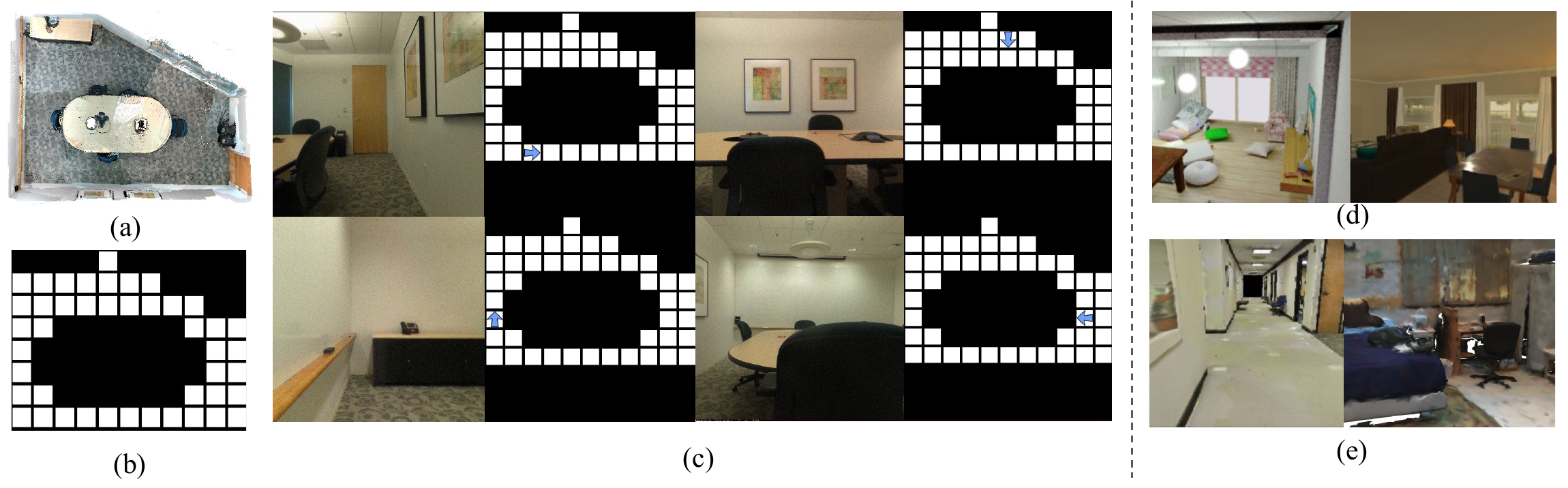}
    \caption{The AdobeIndoorNav Dataset and other 3D scene datasets. Our dataset supports research on robot visual navigation in real-world scenes. It provides visual inputs given a robot position: (a) the original 3D point cloud reconstruction; (b) the densely sampled locations shown on 2D scene map; (c) four examples RGB images captured by robot camera and their corresponding locations and poses. Sample views from 3D synthetic and real-world recontructed scene datasets: (d) Observation images from two synthetic datasets: SceneNet RGB-D \cite{handa2015scenenet, mccormac2016scenenet} and AI2-THOR \cite{zhu2017target, kolve2017ai2}; (e) Rendered images from two real-world scene datasets: Stanford 2D-3D-S \cite{armeni2017joint} and ScanNet \cite{dai2017scannet}.}
    \label{fig:dataset_overview}
\end{figure*}

The simulation of navigation is implemented by rendering visual inputs to a virtual robot moving in the scene. To further reduce the exploration space for DRL, the robot is restricted to walk only on a set of grid locations in the scene. Hence, the datasets need to provide real-world visual inputs at every grid position to the robot. With synthetic scenes, it is straightforward to do the rendering from 3D models in DRL training, as shown in Figure~\ref{fig:dataset_overview} (d). However, since the synthetic scenes do not reflect the real-world scene complexity, they are not ideal to understand how an algorithm works in real-world scenarios. On the other hand, 3D reconstructed scenes contain real-world scene details. However, the rendered visual inputs from point cloud or 3D meshes are usually with holes and artifacts, as shown in Figure~\ref{fig:dataset_overview} (e), due to the low-quality of the 3D reconstructed data. To facilitate the research in DRL for robot navigation, it demands a dataset that provides high-quality real visual inputs at densely sampled locations in real-world scenes. 


\begin{figure}[t]
    \centering
    \includegraphics[width=\linewidth]{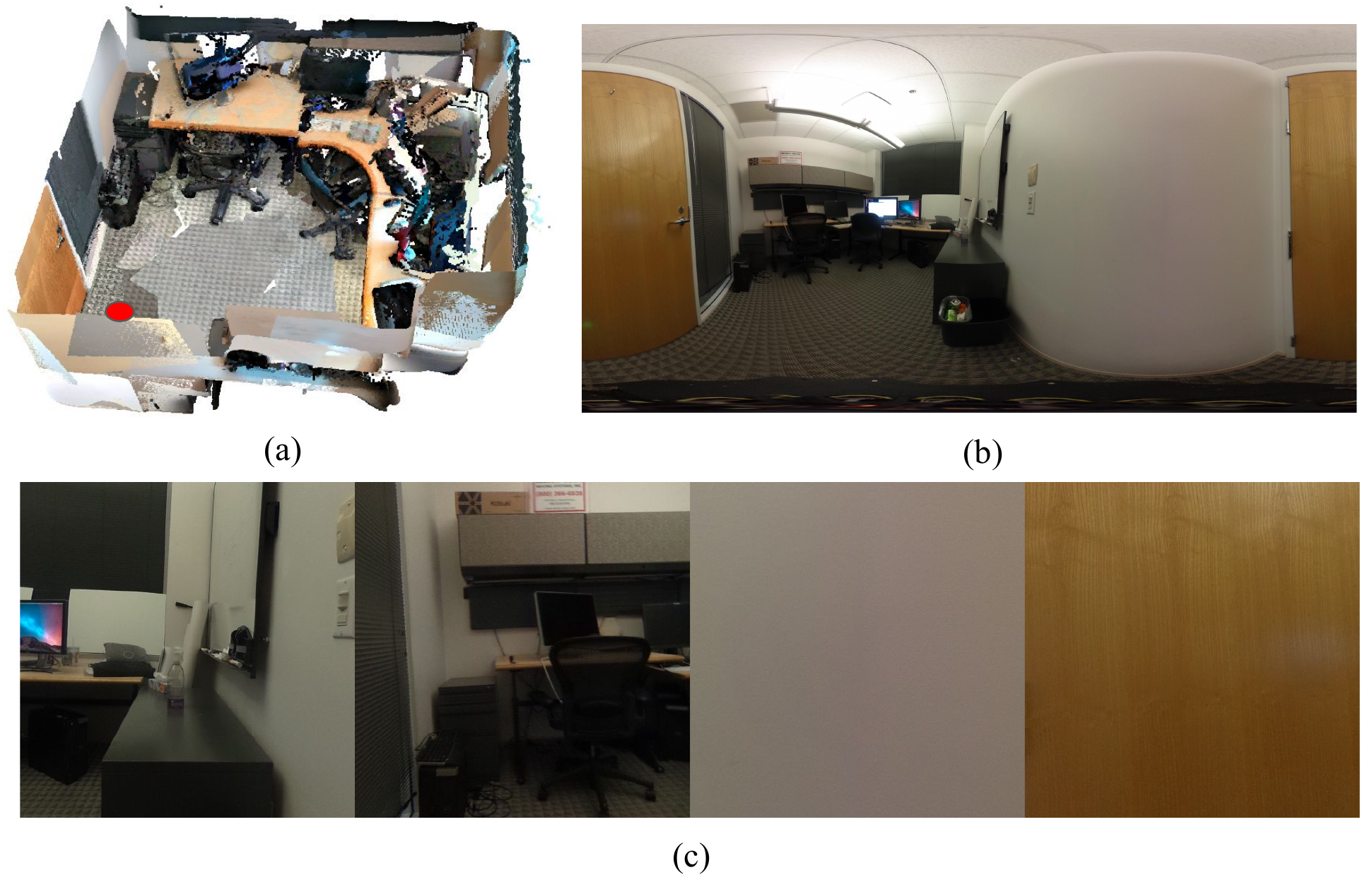}
    \caption{The collected AdobeIndoorNav Dataset: for each scene, we provide a) its 3D reconstruction in point cloud, b) 360-degree panoramic views at densely sampled locations, and c) views in front/back/left/right directions as visual inputs to the robot.}
    \label{fig:dataset_teaser}
\end{figure}

In this work, we propose the AdobeIndoorNav dataset to fill the gap of synthetic 3D scene datasets and 3D reconstructed scene datasets. As shown in Figure~\ref{fig:dataset_teaser}, it contains real-world 3D reconstructed scenes as well as densely captured real-world 2D images. We implement a semi-automatic and easily re-producible pipeline to collect this dataset by taking advantage of low-cost 3D reconstruction solution and SLAM on real robot. Compared with the most recent 3D reconstructed scenes dataset Matterport3D~\cite{chang2017matterport3d} where the grid size is $2.25\pm0.57$ meter on average, our dataset provides more densely sampled grid with size 0.4$\sim$0.5 meter. For example, in a room of $4\times 6$ square meters, the grid size of Matterport3D leads to only 6 locations for robot to navigate while our dataset provides around 100 locations. We will release the dataset to the community.


With the proposed dataset, we study the method by Zhu et al.~\cite{zhu2017target} for DRL based robot visual navigation in real-world scenarios. Zhu et al.~\cite{zhu2017target} models the robot visual navigation problem in a target reaching manner. The robot state is identified by its visual input and the target is specified by its camera view at the target position. In this target-driven visual navigation setting, a DRL algorithm is employed to learn a navigation policy. The deep neural networks determine the desired actuation from the robot’s current state and target state to move the robot in four directions. As shown in Figure~\ref{fig:dataset_overview} (a)-(c), our dataset provides the real-world visual inputs collected at densely sampled positions, which well-supports training this algorithm.

The proposed setting is an intentionally simplified version of real-world robot visual navigation with neither moving obstacles nor continuous actuation. It is interesting and promising to observe that with DRL the robot can learn to navigate to a target state solely based on its current visual input without a-prior constructed map. However, in our experiments, we observe that the weak generalization of the learned policy with their method heavily limits its real-world applicability, even under this simplified setting. In Section~\ref{sec:discussion}, we further discuss our attempts on improving this DRL based method with empirical evaluation.

In summary, the contribution of this paper is in three-fold: (1) we propose a real-world scene dataset for robot visual navigation that fills the gap of 3D synthetic datasets and 3D reconstructed datasets; (2) we describe the detailed pipeline to collect the dataset automated by visual SLAM with low-cost robot platform and sensors; (3) we share our thoughts and attempts towards real-world DRL based robot visual navigation.

\section{Related Works}
\subsection{Indoor Robot Visual Navigation}

Robot visual navigation has a  long history~\cite{giralt1979multi, moravec1980obstacle, nilsson1984shakey, moravec1983stanford, chatila1985position}. People have been investigating vision-based mapping and localization as early as 1980s. The indoor scenarios are especially challenging due to the GPS-denied environment. Existing methods along this line can generally be categorized into two groups: map-based navigation and map-less navigation. With a-prior constructed map, map-based methods allow the robot to plan a path ahead of time and localize itself with respect to the map from visual sensors. Usually the robot maintains a dynamic obstacle map to avoid obstacles when executing the planned path~\cite{moravec1981rover}. When navigating in an environment without a map, the robot can either incrementally build a map while navigating around~\cite{gupta2017cognitive} or simply following a reactive setting to navigate based on current state and target state~\cite{zhu2017target}.

\subsection{Deep Reinforcement Learning}

Ever since the success of seminal Deep Q-Networks (DQN) algorithm on playing Atari games~\cite{mnih2015human} and beating human GO players~\cite{silver2016mastering}, DRL received a lot of attention in the research community. Recent years observed great progresses on applying DRL to robotics as well. To name a few, Kohl et al.~\cite{kohl2004policy} introduce policy gradient RL method for locomotion of a four-legged robot. Peters et al.~\cite{peters2008reinforcement} use RL to learn motor primitives. Silver et al.~\cite{silver2014deterministic} extend the RL methods to handle MDPs with continuous actions. Kahn et al.~\cite{kahn2017uncertainty} introduce uncertainty-aware RL method to learn to navigate an a-priori unknown environment while avoiding collisions with applications on quad rotors and RC cars. The potential of learning a model-free optimal policy from end-to-end with minimal human supervision makes DRL a highly promising direction to pursue.

\subsection{Datasets}

DRL, as a data-driven method, demands a proper dataset to tune its neural networks for indoor robot visual navigation. To collect sufficient amount of trajectories to train the policy, online learning in a real-world scenario on a real robot is still far from realistic at this stage. A feasible and common practice is to leverage a simulator to generate visual inputs to the robot based on its position in a specific scene, which relies on 3D scenes datasets.

Synthetic 3D scenes provide plentiful annotations almost for free and have been widely used in research on scene understanding. Most of them can readily be leveraged to render visual inputs at a given position to enable training DRL for robot navigation. 

SceneNet RGB-D~\cite{handa2015scenenet, mccormac2016scenenet} and AI2-THOR ~\cite{zhu2017target, kolve2017ai2} are two recently proposed synthetic 3D scenes datasets. As shown in Figure~\ref{fig:dataset_overview} (d), the rendered visual inputs are in impressively high quality. Compared with prior works, they provide photo-realistic textures and diversified scenes. While they are closing the domain gap between synthetic textures and real textures, they are not actually captured in real-world scenarios. Even though they may serve as good resources to train DRL policy, they are not desirable in evaluating the learned policy in real-world scenes.

Capturing the real-world scenes in 3D has been well explored in computer vision. Most of the existing datasets captured from real-world scenes contain certain formats of 3D reconstructed scenes, such as point cloud and 3D mesh. To use them to train DRL based robot navigation, the visual inputs can be rendered from the 3D reconstructions.

ScanNet~\cite{dai2017scannet} design an easy-to-use RGB-D capture system to collect 1,500 real-world scene scans, along with many semantic annotations. Stanford 2D-3D-Semantics Dataset~\cite{armeni2017joint} include real-world 3D scenes that expand to the entire building floors. Matterport 3D dataset~\cite{chang2017matterport3d} introduce a large-scale RGB-D dataset containing 90 building-scale scenes and 194,400 RGB-D images. Despite of the large number of scenes in these datasets, when rendering visual inputs from the point clouds or 3D meshes, we observed holes and artifacts as shown in Figure~\ref{fig:dataset_overview} (e). The quality for the rendered images can not match the real ones as if they were seen by the robot at a given location.

The dataset we present in this paper fills the gap between the above two categories. We design a semi-automatic pipeline to collect 360-view images at densely sampled grid locations in totally 24 scenes. To train DRL with our dataset, one can easily find the corresponding 360-view image at a given location to crop out the desired visual inputs to the robot to navigate. The low-cost setting makes the pipeline reproducible. People can use the pipeline to collect more real-world scene data for their purpose. The collected dataset can serve as both training data for DRL and benchmark for real-world robot visual navigation.

\section{The AdobeIndoorNav Dataset}
In this section, we describe our AdobeIndoorNav dataset and a semi-automatic pipeline to collect the dataset with sensors mounted on a turtle robot. First, we manually scan the scenes and reconstruct 3D point clouds with a mobile phone with RGB-D cameras. Then, we take advantage of the state-of-the-art SLAM technique to automate robot to walk on a grid to collect real-world 2D visual observations at densely sampled locations. The current dataset contains 24 scenes in the office setting.


\subsection{Data Acquisition Pipeline}

\begin{figure*}[h]
    \centering
    \includegraphics[width=\linewidth]{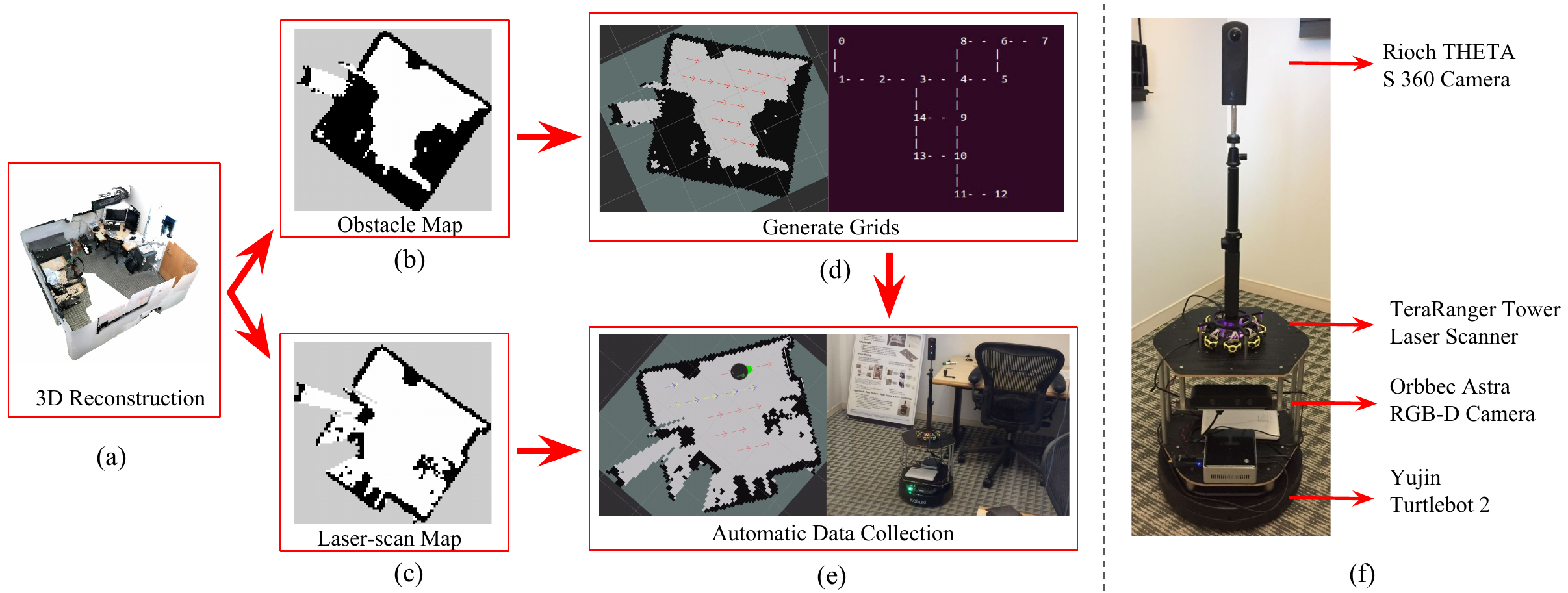}
    \caption{The Pipeline to Collect the AdobeIndoorNav Datatset and the Robot Setting. (a) 3D reconstruction of the scene is obtained by the Tango device; (b) A 2D obstacle map is generated from the 3D point cloud and it indicates the area where robots can navigate; (c) A 2D laser-scan map is generated from the 3D point cloud and it is used later for robot to do localization; (d) Densely sampled grid locations are generated on the 2D obstacle map; (e) Robot runs in the real scenes and captures the RGB-D and panoramic images at all grid locations; (f) Our Turtlebot equipped with one RGB-D camera, one 360 panoramic camera and a series of laser scanners.}
    \label{fig:dataset_pipeline}
\end{figure*}

The detailed pipeline is shown in Figure~\ref{fig:dataset_pipeline}.
First, we manually scan the scene with a Lenovo Phab 2 Tango phone to get the 3D reconstructed scene in point cloud (Figure~\ref{fig:dataset_pipeline} (a)). This process can be readily done using out of the box solution from Tango without prior training. Other 3D reconstructed scenes datasets such as Stanford 2D-3D-S~\cite{armeni2017joint} and Matterport 3D~\cite{chang2017matterport3d}, both use Matterport cameras to do the scanning which require professional operations as well as significant budget. Compared with their acquisition process, our solution is much more portable, lower in cost, and easily reproducible.

Then the 3D reconstructed scene generates two 2D maps: an obstacle map (Figure~\ref{fig:dataset_pipeline} (b)) and a laser-scan map (Figure~\ref{fig:dataset_pipeline} (c)). The obstacle map is generated by aggregating the occupancy maps in 3D within the height range of the robot. The obstacle map identifies the free space for the robot to move around without collision in the entire scene. It is also used later to sample the grid locations to collect visual inputs. The laser-scan map is a 2D map that summarizes the 3D occupancy map at the height of the RGB-D sensor on our robot, which can be leveraged later for localization.

Next, we generate grid locations on the obstacle map with a simple Depth-first Searching (DFS) algorithm. The starting position is specified by the user. We take a grid size of $0.5$m$\times0.5$m in most scenes and reduce it to $0.4$m$\times0.4$m for smaller scenes.

After the manual efforts, the most time-consuming data collection part is automatically done by the robot. The depth channel from the RGB-D sensor is used as a laser scanning to localize itself with respect to the laser-scan map~\cite{bailey2006simultaneous}. Then the robot walks on the grid locations following the generated DFS path to collect data. It stops at each grid location to take the 360-view image. We also have a set of range sensors around to avoid hitting into unexpected obstacles.

\subsection{Dataset Statistics}
The AdobeIndoorNav dataset currently contains 24 indoor rooms: 15 offices, 5 conference rooms, 1 storage room, 1 kitchen and 2 open spaces, all of them are collected inside Adobe buildings. There are totally 3,544 densely sampled locations. On average, each scene has 148 locations for robots to navigate. We split the entire dataset into a train split (15 scenes) and a test split (9 scenes).

\setlength{\tabcolsep}{3pt}

\begin{table}[]
    \centering
    \caption{The AdobeIndoorNav Dataset Statistics}
    \begin{tabular}{l|c|ccc}
    \hline
         \textbf{Scene Name} &	\textbf{Split} & \textbf{Total Locs}&	\textbf{Target Locs}&	\textbf{Featureful Locs}\\
         \hline
et07-imagination-lab& train &	284&	7	&267\\
et12-corner-1&test &	156	&6	&131\\
et12-corner-2&train &	388&	8&	337\\
et07-cr-galgary	&test &164&	8&	99\\
et12-cr-hamburg	&train &176&	6&	111\\
et12-cr-helsinki&train &	168	&6&	105\\
et12-cr-hongkong&train &	260	&8&	175\\
et12-cr-honolulu&test &	252	&7&	149\\
et12-kitchen&test &	424	&10&	396\\
et12-office-104&train &	40	&5&	32\\
et12-office-108&train &	76&	5&	60\\
et12-office-110&train &	60&	5&	52\\
et12-office-111&train &	68&	5&	55\\
et12-office-112&train &	72&	5&	63\\
et12-office-113&train &	96&	5&	61\\
et12-office-114&train &	84&	4&	63\\
et12-office-115&train &	72&	4&	53\\
et12-office-117&train &	68&	5&	39\\
et12-office-132&train &	232&	7&	144\\
et07-office-114&test &	48&	4&	39\\
et07-office-419&test &	68&	5&	57\\
et07-office-420&test &	76&	5&	49\\
et07-office-423&test &	100&	5&	73\\
et07-office-424&test &	112&	6&	81\\
\hline
\textbf{total} & &	3,544&	141&	2,691\\
\hline
    \end{tabular}
    \label{tab:dataset_stats}
\end{table}

\begin{figure}[h]
    \centering
    \includegraphics[width=0.8\linewidth]{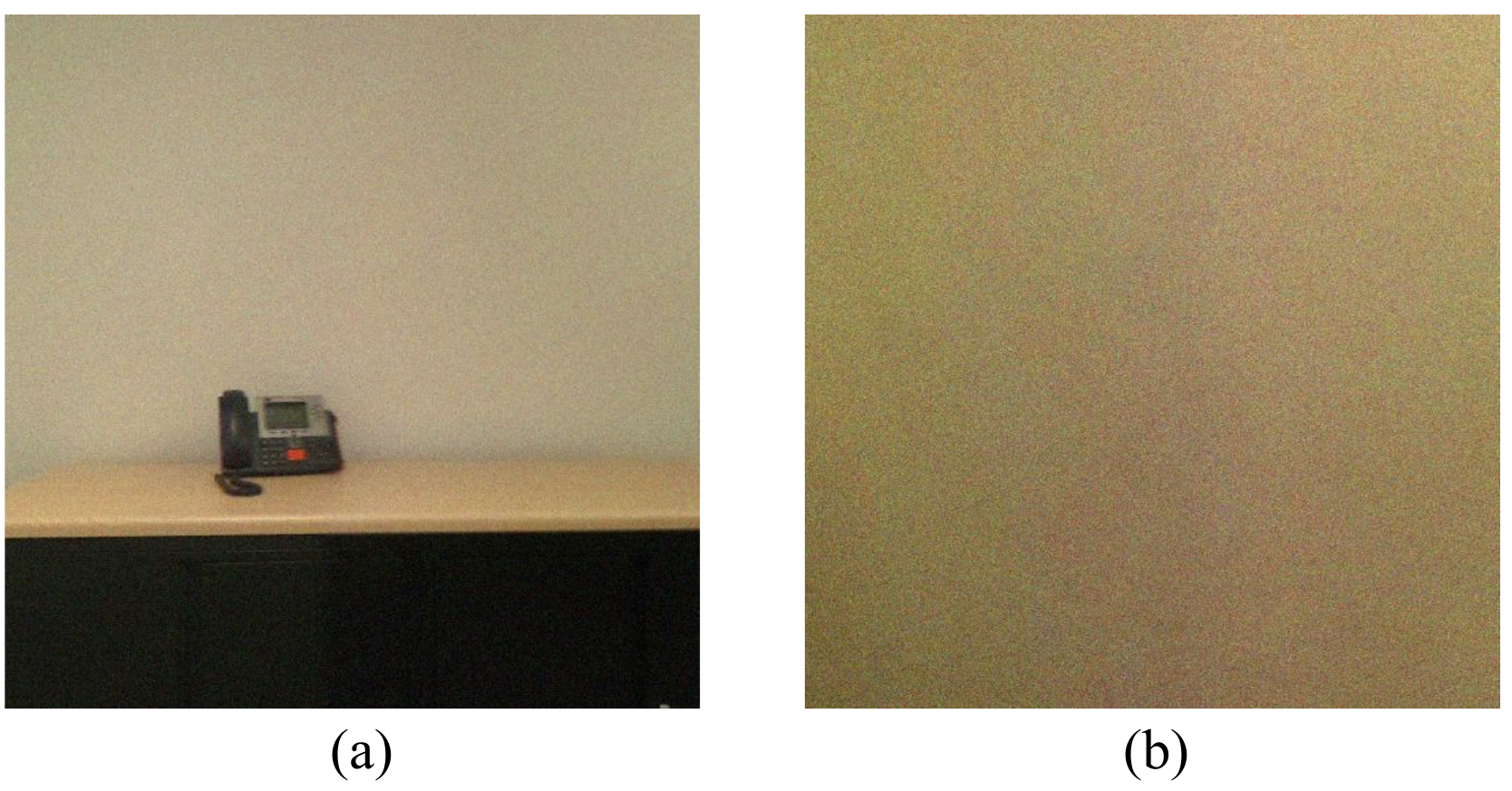}
    \caption{Sample Landmark and Feature-less Image Views. (a) Sample featureful target; (b) Sample feature-less target.}
    \label{fig:dataset_feature_less}
\end{figure}

Table~\ref{tab:dataset_stats} summarizes the statistics on the 24 scenes in our dataset. To provide a benchmark for the community, we identify a set of landmark target locations for each scene (5$\sim$10 per scene), as suggested in \cite{zhu2017target}. We also detect SIFT key points~\cite{lowe2004distinctive} from all visual inputs to identify targets with distinctive image features. Figure~\ref{fig:dataset_feature_less} shows a sample featureful target and a feature-less one. The featureful targets include all interesting targets for navigation and are super-sets of the landmark ones. The \textit{Target Locs} and \textit{Featureful Locs} columns summarize the number of the landmark targets and these featureful targets respectively. 


With the visual inputs collected at all grid locations, it is straightforward to use our dataset to train and test robot visual navigation. In addition to that, with the densely collected 360-view images, our dataset can potentially be used to study view selection. We will release the dataset to public and we hope the dataset can facilitate research in multiple areas.


\section{Experiments}
In this section, we report our experiments on target-driven visual navigation with our dataset. We follow the recently proposed method by Zhu et al.~\cite{zhu2017target} to train navigation policies with DRL. As their method is essentially a A3C model~\cite{mnih2016asynchronous}, we use A3C to refer to their method.

To briefly review their method, Zhu et al.~\cite{zhu2017target} use a Siamese A3C model that takes the most recent four frames of visual observations and a target view as inputs to predict the action to execute. The action is one of the four pre-defined movements including moving forward, moving backward, turning left, and turning right. In training, the robot is given a large goal-reaching reward upon task completion and small step penalty to encourage short trajectory. We evaluate the proposed method and a number of its variants on our dataset. We also include the result from random policy and the ground-truth shortest-path length for comparison. The methods we evaluate are the following:
\begin{itemize}
    \item \textbf{Random}: the agent uniformly chooses a random action to take at every timestep.
    \item \textbf{Shortest-path}: the agent has an oracle that tells it the shortest path from the starting location to the target.
    \item \textbf{A3C Four-frame}: the model uses four history frames of observations as input. This is the design by Zhu et al.~\cite{zhu2017target}.
    \item \textbf{A3C One-frame}: the model only uses the current single frame as input.
    \item \textbf{A3C LSTM}: the model uses a LSTM network to keep the memory for history observations.
\end{itemize}

We report the navigation performance on all scenes (24 scenes) as well as five categories of scenes including the office scenes (15 office rooms), the conference room scenes (5 conference rooms), the open area scenes (\textit{et12-corner-1} and \textit{et12-corner-2}), the kitchen scene (\textit{et12-kitchen}) and the storage room scene (\textit{et07-imagination-lab}).

All the models are implemented in Tensorflow~\cite{abadi2016tensorflow} and trained with 100 CPU threads. Each thread trains for one scene and is assigned one landmark target to train. In each training episode, the agent starts from a random position and tries to navigate to the given target. The episode ends when the robot reaches the target or exceeds 500 steps. All the models are trained for about 40 million frames (total length of trajectories).

In evaluating the learned policy, the maximum episode length is set to 10,000. For the episodes that the robot eventually fails to reach the target, we use 10,000 as the episode length. To avoid agent getting stuck or failing in loops, we add 5\% chance of exploration by executing a random policy. For each navigation target, we randomly select 10 different starting positions and report the average episode lengths. 


\setlength{\tabcolsep}{2.5pt}

\begin{table}[h]
    \centering
    \caption{Evaluation on the AdobeIndoorNav Dataset for training targets: how the learned policy works on the targets observed in training.}
    \begin{tabular}{c|c|ccccc}
    \hline
          &	\textbf{All} & \textbf{Office} & \textbf{Conf}&	\textbf{Open}&	\textbf{Kitchen} & \textbf{Storage}\\
         \hline
         \#scenes&      24   &   15 & 5 & 2 & 1  & 1 \\
         \hline
         Random &    258.50 & 183.80 & 369.38 & 397.83 & 409.88 & 394.73 \\
         Shortest-path &  7.53 & 5.19 & 11.20 & 11.37 & 15.60 & 8.56\\
         \hline
         A3C One-frame&   56.88 & 6.14 & 14.12 & 586.02 & 17.21 & 13.23 \\
         A3C Four-frame&    35.31 & 6.04 & 14.60 & 303.83 & 19.13 & 56.94  \\
         A3C LSTM &   9.23 & 6.19 & 13.26 & 14.85 & 20.19 & 12.44 \\
         \hline
    \end{tabular}
    \label{tab:eval_on_train}
\end{table}

As shown in Table~\ref{tab:eval_on_train},
the target-driven A3C models demonstrate successful navigation to the targets seen during training. Giving the robot more history frames seems to be helpful to learn a better policy to navigate to training targets.

To evaluate the target generalization of these models, we randomly select 5$\sim$20 featureful targets that the robot has never seen during training and we run the learned policy to navigate to these new targets. Table~\ref{tab:eval_on_test} shows the average episode length for the testing targets. Compared to the success on navigating to the training targets, all the A3C models fail to generalize to new targets. The performance is even worse than the random policy. 

\begin{table}[h]
    \centering
    \caption{Evaluation on the AdobeIndoorNav Dataset for testing targets: how the learned policy generalize to unseen targets.}
    \begin{tabular}{c|c|ccccc}
    \hline
          &	\textbf{All} & \textbf{Office} & \textbf{Conf}&	\textbf{Open}&	\textbf{Kitchen} & \textbf{Storage}\\
         \hline
         \#scenes&      24   &   15 & 5 & 2 & 1  & 1 \\
         \hline
         Random &      256.05 & 187.34 & 370.89 & 371.08 & 399.19 & 339.19 \\
         Shortest-path &   7.37 & 5.37 & 11.00 & 9.97 & 13.23 & 7.99 \\
         \hline
         A3C One-frame&   5543.02 & 4861.32 & 7198.17 & 6100.12 & 7278.01 & 4643.45 \\
         A3C Four-frame&  4468.67 & 3630.33 & 6296.34 & 4751.94 & 6832.84 & 4974.86  \\
         A3C LSTM &  4390.89 & 3843.10 & 5315.40 & 4785.22 & 7552.55 & 4034.85\\
         \hline
    \end{tabular}
    \label{tab:eval_on_test}
\end{table}

Besides its weak target generalization, these A3C methods are weak in generalizing across scenes as well. As reported by Zhu et al.~\cite{zhu2017target}, fine-tuning a learned policy to a new scene requires 10 million frames, which is far from practical in real-world scenario. 

We test the scene generalization issue on our dataset. We train a A3C Four-frame navigation network on 15 train scenes for around 30 million frames and then fine-tune the network to 4 exemplar test scenes: two office rooms (\textit{et07-office-114} with 48 locations and \textit{et07-office-424} with 112 locations) and two conference rooms (\textit{et07-cr-galgary} with 164 locations and \textit{et12-cr-hongkong} with 260 locations). For each test scene, we train a new scene-specific layer from scratch. 

We observe that the network converges after fine-tuning for 3 million frames, which takes about six hours on our machine with 40 Intel-i7 CPUs.  Table~\ref{tab:eval_on_test_finetune} reports the navigation performance evaluated on the landmark targets for the four test scenes (\textbf{Finetune}). For comparison, we also include the results of a model trained from scratch on the entire 24 scenes for 40 million frames (\textbf{Scratch}), a model trained with unified scene layer on 15 train scenes and directly tested on the four test scenes (\textbf{Unified}), the random policy (\textbf{Random}) and the shortest-path lengths (\textbf{Shortest}). 

\setlength{\tabcolsep}{1pt}

\begin{table}[h]
    \centering
    \caption{Evaluation of Scene Generalization: how the learned policy fine-tune to unseen scenes.}
    \begin{tabular}{c|cccc}
    \hline
          & et07-cr-galgary	& et12-cr-hongkong	& et07-office-114	& et07-office-424 \\
         \hline
         Random & 376.39 & 359.94 & 106.63 & 179.08 \\
         Shortest & 10.98 & 12.50 & 3.60 & 5.07\\
         Unified&1697.96&2514.66&97.92&1385.70\\
         \hline
         Scratch&14.64&16.18&4.73&6.83\\
         Finetune&15.19 &  2515.44 &   5.10 &   5.97 \\
         \hline
    \end{tabular}
    \label{tab:eval_on_test_finetune}
\end{table}

Even though 3 million frames of fine-tuning successfully adapts the learned policy to the small scenes (i.e. \textit{et07-cr-galgary}, \textit{et07-office-114} and \textit{et07-office-424}), it fails to generalize to large scenes (i.e. \textit{et12-cr-hongkong}). Also, given the fact that training a model on 24 scenes from scratch only takes about 10 million frames to converge, as illustrated by the blue curve in Figure~\ref{fig:exp_dense_train_curve}, 3 millions frames of fine-tuning is evidently heavy.

\section{Discussion~\label{sec:discussion}}
As we observe in the experiments, although DRL for robot visual navigation has its potential, the current technology is still in its infancy. In this section, we report our efforts on improving it as well as attempts and thoughts on future directions.

\subsection{Dense-target Training}

Despite of its exceptional capability to navigate to the training targets, the A3C models fail dramatically on unseen testing targets. This over-fitting problem suggests that the agent only learns to memorize the right action to take at certain location given one target, instead of understanding the spatial relationship from current state to the target and the consequence of taking one of the four actions.
We argue that the issue comes from the fact that only 5$\sim$10 sparsely spreading targets per scene were used in training in~\cite{zhu2017target}. To avoid over-fitting and encourage the agent to learn more generalizable knowledge, we need to let the agent work on more targets in training.

To learn a navigation model that generalizes to all targets in the scenes, we propose to densely sample training targets and train the navigation A3C models on all of them. 

\begin{figure}[h]
    \centering
    \includegraphics[width=0.8\linewidth]{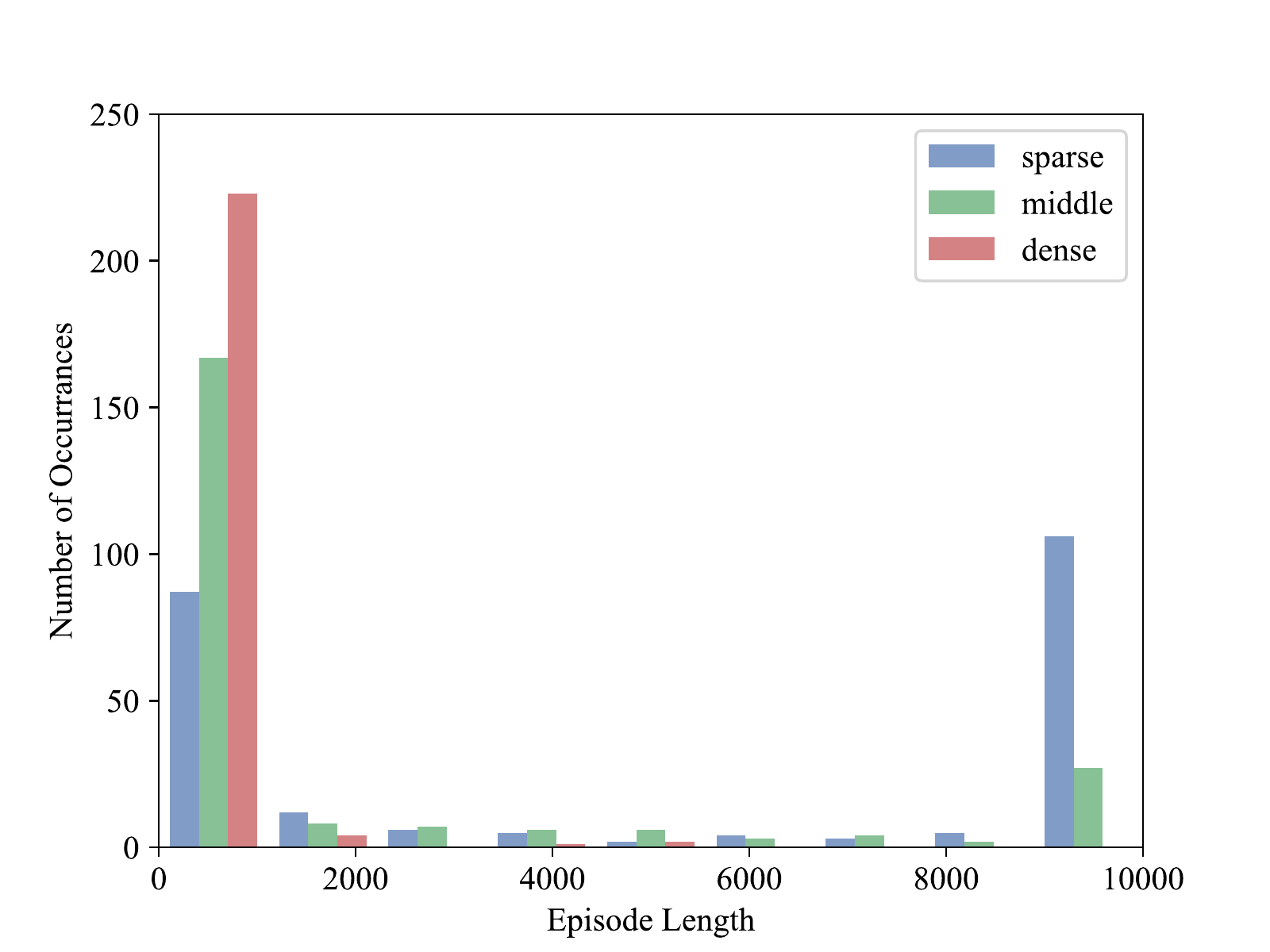}
    \caption{Episode Length Statistics for the Testing Targets in the \textit{et12-kitchen} Scene. We report the evaluation of the models that are trained on 10 (sparse), 100 (middle), and 373 (dense) samples. We run 10 episodes with different random starting locations for each one of 23 testing targets. The maximum episode length is set to 10,000.}
    \label{fig:exp_kitchen_dense_train}
\end{figure}

\begin{figure}[h]
    \centering
    \includegraphics[width=0.8\linewidth]{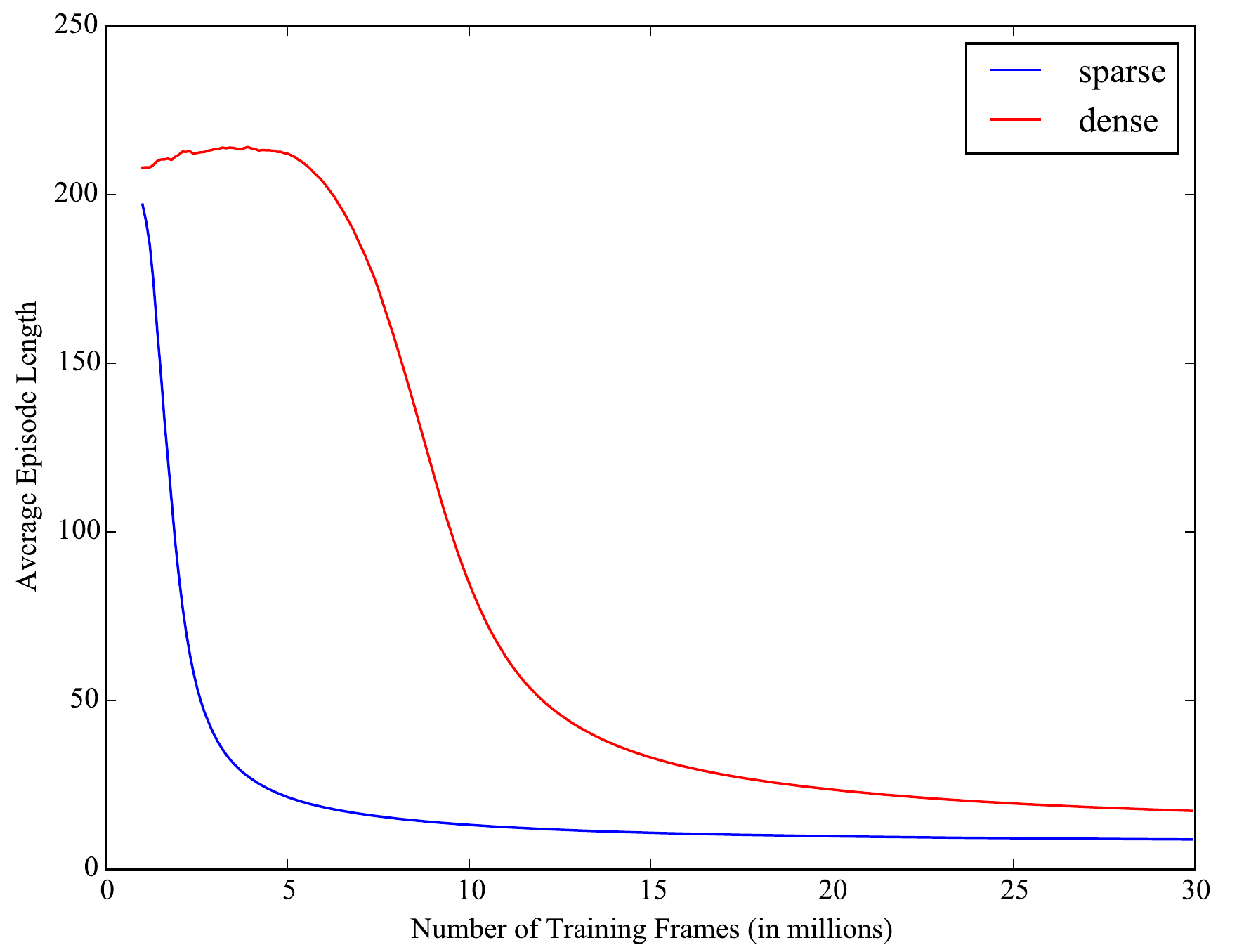}
    \caption{The A3C Model Training Progress Curves: dense-target training takes longer to converge.}
    \label{fig:exp_dense_train_curve}
\end{figure}

We validate this dense target training idea on the \textit{et12-kitchen} scene, which is the largest scene in our dataset. We randomly select 23 out of the 396 distinctive views for testing. We avoid using texture-less views as targets, since they can introduce confusion in learning. Except for the selected testing targets, all the rest are for training. We train three A3C models with different number of training targets: (a) trained on 10 sparsely sampled targets, (b) trained on 100 sampled targets, and (c) trained on the rest 373 locations. We train all the three models for about 20 million frames. Figure~ \ref{fig:exp_kitchen_dense_train} shows the episode length statistics. In general, when trained with more targets, the robot navigates better at testing targets. 10 sparsely sampled training targets end up with most failure episodes. Similar behavoirs have been observed on other scenes as reported in Table~\ref{tab:exp_dense_train_table}. However, the gain is at the cost of longer training time as shown in Figure~\ref{fig:exp_dense_train_curve}.

\setlength{\tabcolsep}{2.5pt}

\begin{table}[h]
    \centering
    \caption{Evaluation of Dense Target Training}
    \begin{tabular}{c|c|ccccc}
    \hline
          &	\textbf{All} & \textbf{Office} & \textbf{Conf}&	\textbf{Open} & \textbf{Storage}\\
         \hline
         \#scenes&      15   &   10 & 3 & 1 & 1  \\
         \hline
         Random &   249.56 & 190.21 & 366.99 & 401.21 & 339.19   \\
         Shortest-path & 7.18 & 5.44 & 11.33 & 11.27 & 7.99  \\
         \hline
         Train on Sparse Targets &  4359.36 & 3841.52 & 5409.82 & 5770.88 & 4974.86 \\
         Train on Dense Targets &  197.92 & 7.10 & 487.56 & 1401.07 & 34.10 \\
         \hline
    \end{tabular}
    \label{tab:exp_dense_train_table}
\end{table}


\subsection{Spatial-aware Feature}

\begin{figure}[h]
    \centering
    \includegraphics[width=\linewidth]{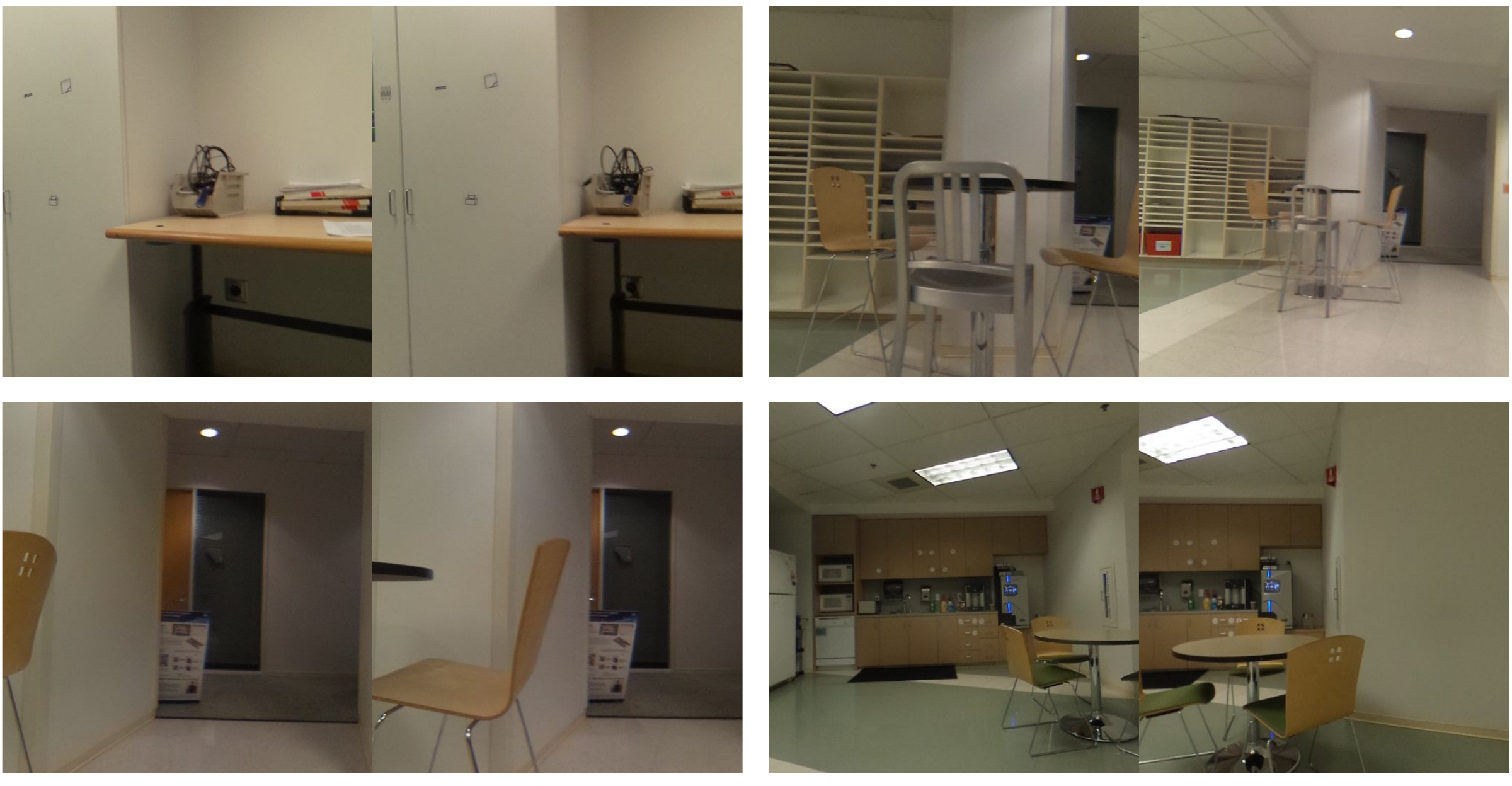}
    \caption{Near target failure cases: In each image pair, the left one shows the current observation and the right image is the target view. In all these examples, the agent fails to reach the target even though the target is just nearby.}
    \label{fig:method_spatial_aware_fail}
\end{figure}

By carefully examining the failure cases, we observed a lot of near target failures as shown in Figure~\ref{fig:method_spatial_aware_fail}. Although the robot reaches a nearby location to the target, it fails to output the right action to move to the target.

A very common scenario in testing is that after starting from a random location, the agent quickly navigates to the area around the testing target in a few steps. This behavior shows that the model successfully learns to navigate to this target probably by learning a rough room layout. However, after reaching the nearby area, the agent begins to vacillate around the target location and finally fails to arrive at the exact target view. It suggests that the robot can not differentiate two nearby and similar views.

To address this issue, we propose to use spatial-aware feature to represent the visual inputs. Instead of using the 2048-dim feature extracted from the last layer of a pre-trained CNN~\cite{he2016deep}, we can leverage features extracted from early layers which encode more spatial information.

The 2,048-dim feature lacks the spatial information on the image space because of the precedent global average pooling operation. We choose to use the $7\times7\times2,048$ feature maps right before the last global average pooling layer to as the representation to address this issue.



\begin{figure}[h]
    \centering
    \includegraphics[width=\linewidth]{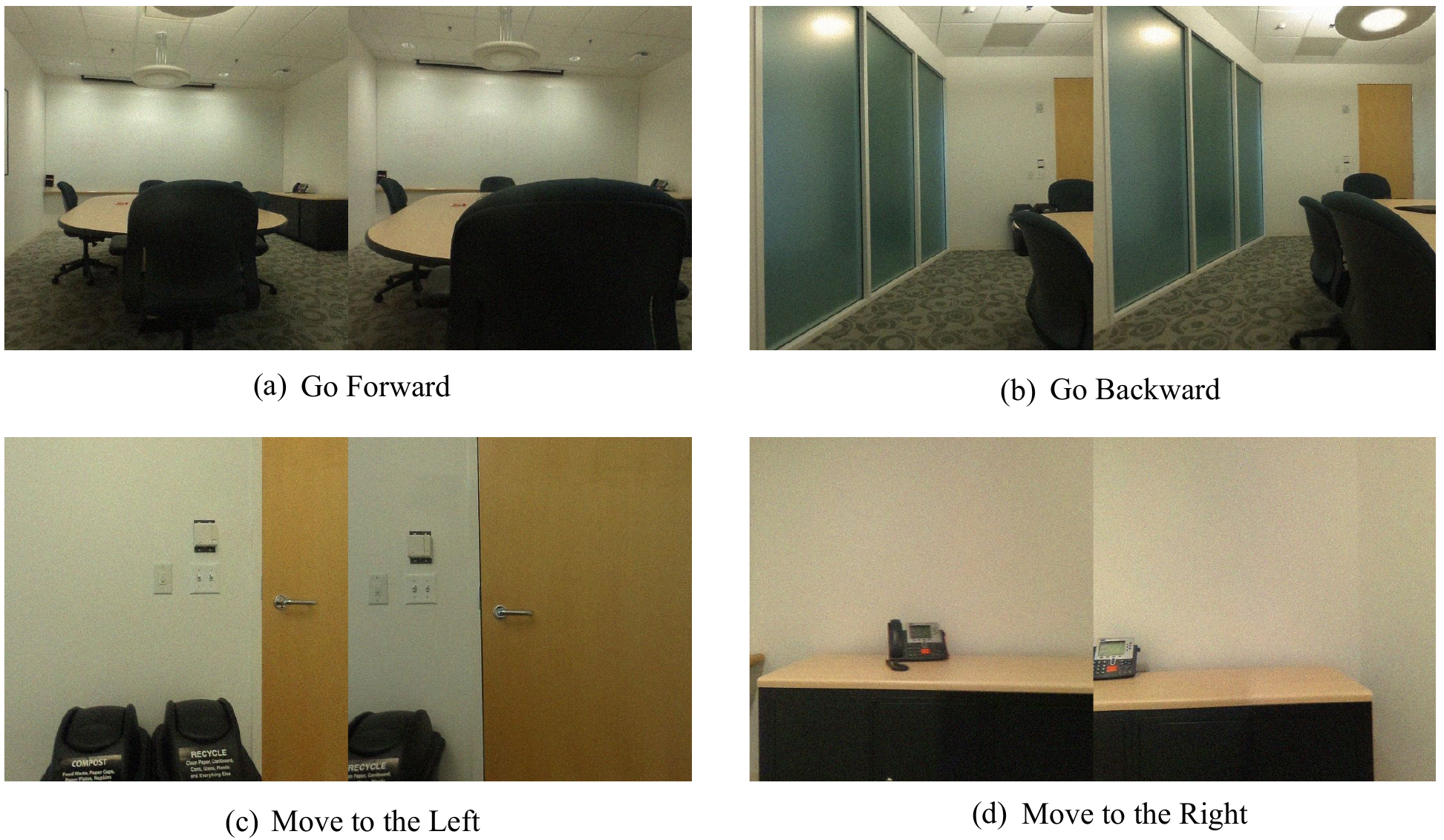}
    \caption{The four types of near-by target testing: the left image shows the current view and the right image shows the target view, which can be reached by (a) going forward; (b) going backward; (c) moving to the left; and (d) moving to the right.}
    \label{fig:exp_spatial_aware_examples}
\end{figure}

We propose a diagnostic experiment to validate this design. Given a pair of images that have a large portion of view overlap and are only one-step away (i.e. go forward, go backward, move to the left, move to the right), we train a simple siamese-style network over extracted image features to predict the action to match the two views. The four types of testing image pairs are illustrated in Figure~\ref{fig:exp_spatial_aware_examples}.

We first train two models using 2,048-dim input features and the proposed spatial-aware features on the image pairs from all 24 scenes in the AdobeIndoorNav dataset. The image pairs for training and testing are different, but are all taken from the same scenes. In Table \ref{tab:diag_spatial_aware}, we observe a classification accuracy improvement from 83.5\% to 87.5\% when we replace the original 2,048-dim feature to the spatial-aware feature. Then, we train the two models on image pairs from the 15 training scenes and test on the other 9 scenes for testing. We observe larger performance improvement from 53.0\% to 72.6\%. This experiment validates that the spatial-aware feature encodes more spatial information that can be used to resolve the near-by target failures.

\begin{table}[h]
    \centering
    \caption{Diagnostic Experiments on Spatial-aware Feature}
    \begin{tabular}{c|cc}
    \hline
          & \textbf{2,048-dim Feature} & \textbf{Spatial-aware Feature} \\
         \hline
         On the same scenes &   83.5\% & 87.5\% \\
         On different scenes &  53.0\% & 72.6\% \\
         \hline
    \end{tabular}
    \label{tab:diag_spatial_aware}
\end{table}

\subsection{Future Work}

The target-driven visual navigation setting is very difficult even for human. Imagine one being dropped in a new environment and given an image of a target view. It is largely impossible even for human to figure out the shortest path. We believe one promising future direction could be combing map-building methods with DRL to construct a map either implicitly or explicitly for navigation. It is not clear how to formulate everything into a DRL framework, but we believe it is an interesting future work.

Another direction to explore DRL for robot navigation is to design a task simpler than the target-driven setting. For example, a more intuitively feasible task could be unknown environment exploration. In this case, DRL can be trained more efficiently since there is a fixed goal independent to the scenes.

Navigation target can also be specified as a relative position to the robot. This paradigm would be easier to address too. The relative position implicitly suggests a shortest path, which would serve as a strong regularization in DRL training.

Despite of the problems we observed by evaluating this DRL based visual navigation algorithm over our dataset, we still believe DRL is promising towards real-world visual navigation.



\section{CONCLUSIONS}
We collect a dataset for the community to study indoor robot visual navigation with deep reinforcement learning. The collected dataset is a simplified version of real-world robot visual navigation. To support training DRL for robot navigation, our dataset fill-in the gap between synthetic 3D scenes datasets and 3D reconstructed datasets. We provide 360-degree panoramic images at densely sample grid locations in 24 scenes. We further study the recently proposed DRL algorithm for the target-driven visual navigation problem in real-world scenes with our dataset. Our observation is that the DRL based method is still far from practical. We discuss our proposed improvements to the method with empirical validation and share our thoughts on potential future work in this direction.

\addtolength{\textheight}{-8cm}   



\bibliographystyle{IEEEtran}
\bibliography{IEEEabrv,IEEEexample}

\end{document}